\documentclass[jimaging,article,accept,pdftex,moreauthors]{Definitions/mdpi} 
\usepackage{bm}
\firstpage{1} 
\pubvolume{1}
\issuenum{1}
\articlenumber{0}
\pubyear{2023}
\copyrightyear{2023}
\externaleditor{Academic Editor: Donald Bailey}
\datereceived{28 August 2023} 
\daterevised{24 October 2023} 
\dateaccepted{31 October 2023} 
\datepublished{3 November 2023} 
\hreflink{https://doi.org/} 

\Title{OW-SLR: Overlapping Windows on Semi-Local Region for Image Super-Resolution}

\TitleCitation{OW-SLR: Overlapping Windows on Semi-Local Region for Image Super-Resolution}

\Author{{Rishav Bhardwaj} 
 $^{1,}$*\orcidA{}, Janarthanam Jothi Balaji $^{2}$\orcidB{} and Vasudevan Lakshminarayanan $^{1}$\orcidC{}}

\AuthorNames{Rishav Bhardwaj, Janarthanam Jothi Balaji and Vasudevan Lakshminarayanan}

\AuthorCitation{Bhardwaj, R.; Balaji, J.J.; Lakshminarayanan, V.}

\address{%
$^{1}$ \quad School of Optometry and Vision Science, University of Waterloo, {Waterloo, ON N2L 3G1}
, Canada; vengulak@uwaterloo.ca \\
$^{2}$ \quad Department of Optometry, Medical Research Foundation, Chennai {603103}, India; jothibalaji@gmail.com}

\corres{Correspondence: rishav.bhardwaj@uwaterloo.ca}

\abstract{There has been considerable progress in implicit neural representation to upscale an image to any arbitrary resolution. However, existing methods are based on defining a function to predict the Red, Green and Blue (RGB) value from just four specific loci. Relying on just four loci is insufficient as it leads to losing fine details from the neighboring region(s). We show that by taking into account the semi-local region leads to an improvement in performance. In this paper, we propose applying a new technique called Overlapping Windows on Semi-Local Region (OW-SLR) to an image to obtain any arbitrary resolution by taking the coordinates of the semi-local region around a point in the latent space. This extracted detail is used to predict the RGB value of a point. We illustrate the technique by applying the algorithm to the Optical Coherence Tomography-Angiography (OCT-A) images and show that it can upscale them to  random resolution. This technique outperforms the existing state-of-the-art methods when applied to the OCT500 dataset. OW-SLR provides better results for classifying healthy and diseased retinal images such as diabetic retinopathy and normals from the given set of OCT-A images. The project page is available at \url{https://rishavbb.github.io/ow-slr/index.html}.
  }


\begin{document}




\section{Introduction}

The primary objective of super resolution (SR) is to obtain a credible high resolution (HR) image from a low resolution (LR) image. The major challenge is to retrieve the information which is too minute or almost non existent, and to extrapolate this information to higher dimensions which is plausible to the human eye. Furthermore, the availability of paired HR-LR image data poses another concern. Typically, an image is downsampled using a specific method in the hope of encountering a real-life LR image that is somewhat similar. The aim of SR models is to fill in the deficient information between the HR and LR images, thereby bridging the gap. Also, for high-dimensional inputs like videos and 3D scans there are quite a few work in the literature \cite{bashir2021comp, tan2022, li2021spatio, thawakar2019image, li2022srdiff, girshick2016region}.

Most of the architectures \cite{liang2021swinir,shi2016real, zhang2018image, peng2020convolutional, lim2017enhanced} proposed for SR of images upsample them by a fixed factor only. This means that a separate architecture needs to be trained for each unseen upscaling factor. However, the real world is continuous in nature, whereas images are represented and stored as discrete values in 2D arrays. Inspired by \cite{saito2019pifu, park2019deepsdf, mescheder2019occupancy, sitzmann2020implicit} for 3D shape reconstruction using implicit neural representation, {ref.} 
 \cite{chen2021learning} proposed Local Implicit Image Function (LIIF) to represent images in a continuous fashion. Some postprocessing is performed to obtain the RGB value of the query point. This approach enables representing and manipulating images in a continuous manner, departing from the traditional discrete representation in 2D arrays.

In our work, we draw partial inspiration from advancements in 3D shape reconstruction, but we extend the approach by considering a semi-local region rather than relying solely on four specific locations. Our method allows for extrapolation to any random upscaling factor using the same architecture.~This architecture takes into account the semi-local region and specifically learns to extract important details related to a query point in the latent space that needs to be upscaled. In this paper, we propose an image representation technique called Overlapping Windows for Semi-Local Representation in a continuous domain and we fine our work as follows: (i) Each image is represented as a set of latent codes, establishing a continuous nature. To determine the RGB value of a point in the HR image within the latent space, we employ a decoding function. (ii) This semi-local region is fed into network as input which generates the embeddings of the intricate details in it which have high probability of getting lost when an entire image is taken into consideration by the networks. (iii) The overlapping window technique allows for effective learning of features within the semi-local region around a point in the latent space using the embeddings. (iv) A decoder takes the features derived from the overlapping window technique and produces the RGB value of the corresponding point in the HR image.

In summary, our work makes two key contributions. Firstly, we introduce a novel technique called overlapping windows, which enables efficient learning of features within the semi-local region around a point. This approach allows for more effective representation and extraction of important details. Secondly, our architecture is capable of upscaling an image to any arbitrary factor, providing flexibility and versatility without the need for separate architectures for different upscaling factors. This contribution enables seamless and consistent image upscaling using a unified framework.


\section{Related Work}
During the early stages of SR research, images were typically upsampled by a certain factor using simple interpolation techniques, and the network was trained to learn the extrapolation of the LR images \cite{dong2015image, kim2016accurate}. However, this approach presents some issues. Firstly, the pre-upsampling process introduces more parameters compared to the post-upsampling process. Pre-upsampling is defined as upscaling the input image and then passing it through the network, whereas post-upsampling is defined as passing the image through the network and then upscaling the feature map. Secondly, due to the higher requirement of parameters more training time becomes a requisite. The network needed to learn the intricacies of the pre-upsampling method, which added to the overall training complexity. Finally, the pre-upsampling process using traditional bicubic interpolation does not yield realistic results during testing. Since it is the first step of the SR pipeline, the network often attempts to mimic this interpolation, which limits the realism of the output images. On the other hand, post-upsampling approaches, where the LR image is downscaled in the very first step, typically involve the use of bicubic interpolation for resizing. However, downscaling an image, even with bicubic interpolation, tends to yield more realistic results compared to upscaling.
As a result, the research focus has shifted towards post-upsampling techniques, which provides more efficient and realistic SR results by leveraging downscaling with appropriate interpolation methods in the very first step.

As already mentioned, downscaling of images happens as the initial step in post-upsampling process. The network learns features from the downscaled image and the upsamples the learned features towards the very end. A technique proposed by Shi et~al. in their work \cite{shi2016real} is known as sub-pixel convolution. Sub-pixel convolution handles the extrapolation of each pixel by accumulating the features along the channel of that pixel. By rearranging the feature channels, sub-pixel convolution enables the network to effectively upscale the LR image to a higher resolution. While sub-pixel convolution provides a practical solution for upsampling by integral factors ($\times1, \times2, \times3$, etc.), it does not support fractional upsampling factors ($\times1.4, \times2.9$, etc.). However, for cases where fixed integral upsampling factors are sufficient, sub-pixel convolution offers an efficient approach to achieving high-quality upsampling. The work by Ledig et~al. \cite{ledig2017photo} introduced the use of multiple residual blocks for feature extraction in super-resolution (SR) tasks. Their approach demonstrated the effectiveness of residual blocks in capturing and enhancing image details. Building upon Ledig et~al.'s work, Lim et~al. \cite{lim2017enhanced} proposed an enhanced SR model that incorporated insights regarding batch normalization. They postulated that removing batch normalization from the residual blocks could lead to improved performance for SR tasks. This is because batch normalization tends to normalize the input, which may reduce the network's ability to capture and amplify the fine details required for SR. Removing batch normalization not only results in a reduction in memory requirements but also makes the network faster. Additionally, the work by Shi et~al. \cite{shi2016real} contributed to the development of various approaches for SR using CNNs. These approaches include methods proposed by \cite{zhang2018image,ledig2017photo,mei2021image,zhang2018residual}. These methods aimed to enhance feature extraction capabilities specifically tailored for SR problems, further advancing the state-of-the-art in SR research.

After the success of CNNs in SR tasks, researchers explored the use of generative adversarial networks (GANs) to further improve SR performance. Several works, such as~\cite{ledig2017photo, sajjadi2017enhancenet, wang2018recovering}, introduced different GAN architectures for extrapolating low-resolution (LR) images to higher resolution. ESRGAN (Enhanced Super-Resolution Generative Adversarial Network) proposed by Wang et~al. \cite{wang2018esrgan} introduced a perceptual loss function and modified the generator network to produce HR images. This perceptual loss function aimed to align the visual quality of the generated HR images with that of the ground truth HR images, improving the perceptual realism of the results.

In Real-ESRGAN \cite{wang2021real}, the authors addressed the issue of using LR images downsampled with simple techniques like bicubic interpolation during training. They note that real-world LR images undergo various types of degradations, compressions, and noise, unlike the simple interpolation-based downsampling. To simulate realistic LR images during training, they proposed a novel technique that subjected the training images to various degradation processes, mimicking real-life scenarios. Additionally, Real-ESRGAN introduced an U-Net discriminator to enhance the adversarial training process and improve the quality of the generated HR images.


\section{Method}
\begin{figure}[H]
\begin{adjustwidth}{-\extralength}{0cm}
\centering
\includegraphics[width=17.5cm]{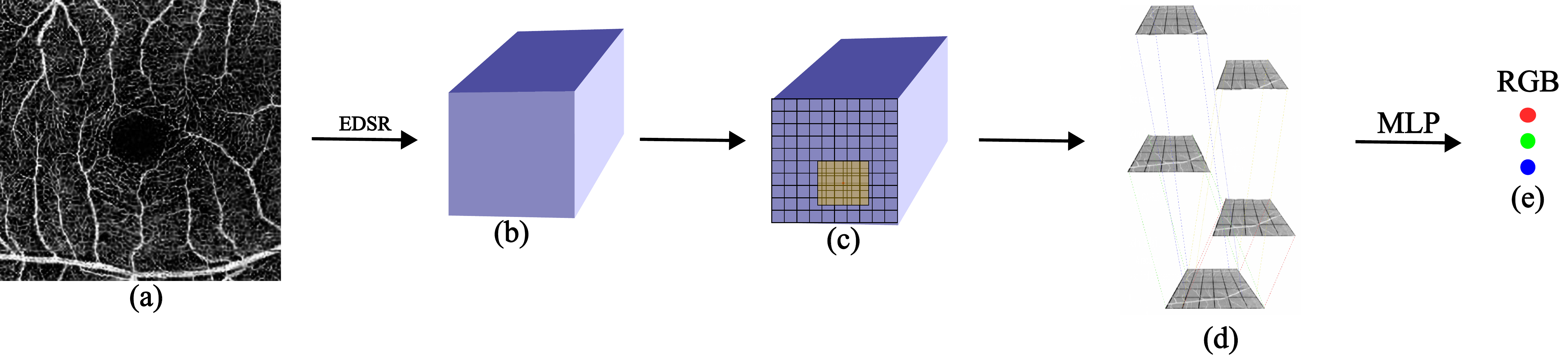}
\end{adjustwidth}
\caption{(\textbf{a}) {An} 
 LR image is taken. (\textbf{b}) It is passed through EDSR~\cite{lim2017enhanced} and a feature map is produced. (\textbf{c}) Locating the semi-local region (M~=~6) around a random selected point from HR image.  (\textbf{d})~Semi-local region is passed through the proposed Overlapping Windows.  (\textbf{e}) This output is passed through the MLP to give out the RGB value of a randomly selected point. Steps (\textbf{c}{--}
\textbf{e}) are performed for all the points in the HR image.\label{fig0}}
\end{figure}

We illustrate the three main components of our approach in this section along with its pictorial representation in Figure~\ref{fig0}. In Section~\ref{sec3.1}, we introduce the backbone of our framework. We represent the LR image as a feature map, which serves as the basis for subsequent processing and analysis. In Section~\ref{sec3.2}, we demonstrate how we find the semi-local region of an arbitrary point in the HR image. This region contains valuable information that helps determine the corresponding RGB value. In Section~\ref{sec3.3}, we highlight the Overlapping Windows technique, which plays a crucial role in predicting the RGB value of a point in the HR image. We accomplish this by leveraging the semi-local region extracted around the sampling points of the feature map. These three parts collectively form the foundation of our approach, allowing for accurate prediction of RGB values.
   
\subsection{Backbone Framework\label{sec3.1}}
To extract features from the LR image, we employ the enhanced deep residual networks (EDSR) \cite{lim2017enhanced}. Specifically, we utilize the baseline architecture of EDSR, which consists of 16 residual blocks.
\begin{equation}
 \psi = EDSR(I_{LR})
\end{equation}
Given an LR image denoted as $ I_{LR} \in \mathbb{R}^{H \times W \times C}\  $, we express it in the form of a feature map  $ \psi \in \mathbb{R}^{P \times Q \times D}\  $. 
Here, $H$ and $W$ represent the height and width of the LR image, respectively, and $C$ signifies the number of channels. $P$ and $Q$ represent the spatial dimensions of the feature map, and $D$ denotes the depth of the feature map.

\subsection{Locating the Semi-Local Region\label{sec3.2}}
In our scenario, we aim to predict the RGB value at any random point in a continuous HR image of arbitrary dimensions. Let $I_{HR} \in \mathbb{R}^{X \times Y \times C}$ represent the HR image. To predict the RGB value at a specific point, we first select a point of interest. Then, we identify its corresponding spatially equivalent point in the feature map $\psi$ obtained from the LR image using bilinear interpolation denoted as $\mathbb{f}_{BI}$. 
\begin{equation}
\bm{\hat{x}} = \mathbb{f}_{BI}(x, \psi)
\end{equation}
where $\bm{\hat{x}}$ and $x$ are the 2D coordinates of the $\psi$ and  $I_{HR}$ respectively.

\begin{figure}[H]
\includegraphics[width=7cm]{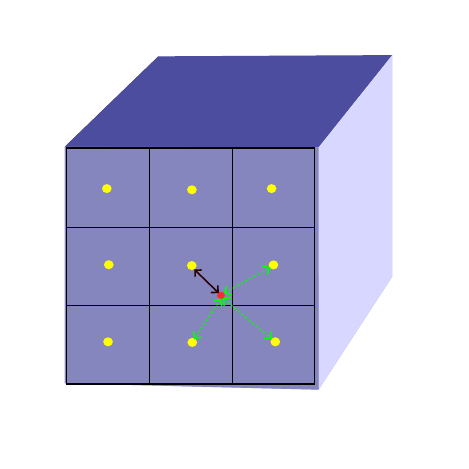}
\caption{To extract features from a feature map of size $3 \times 3$, we focus on a specific query point represented by a red dot. In order to determine which pixel locations in the feature map correspond to this query point, we compute the Euclidean distance between the query point and the center points of each pixel location. In the provided image, the black line represents the closest pixel location in the feature map to the query point.  \label{fig2}}
\end{figure}

Furthermore, we extract a square semi-local region around this corresponding point. The size of this region is determined by a length parameter $M$ units, where each unit dimension of the square region corresponds to the inverse of the dimensions $P$ and $Q$ of the feature map $\psi$ along its length and breadth respectively defined in Equation (5) which is used to find the discrete positions in the semi-local region. Once we have identified the square semi-local region around the corresponding point in the feature map $\psi$, we proceed to extract $M$ $\mathbb{\times}$ $M$ depth features from this region using Equation (3). These depth features capture the important information necessary for predicting the RGB value at the desired point in the HR image. To extract these features, we employ a closest Euclidean distance approach denoted by $\mathbb{g}_{ED}$. Each point within the $M$ $\mathbb{\times}$ $M$ region in $\psi$ is mapped to the nearest point in the latent space, which represents the extracted depth feature. {Figure}  \ref{fig2} illustrates the working of selecting of features from the feature map. This mapping ensures that we capture the most relevant information from the semi-local region.
\begin{equation}
\bm{\hat{X}} = (\bm{\hat{x}}_x - \psi_x * i, \bm{\hat{x}}_y - \psi_y * j)
\end{equation}\vspace{-10pt}
\begin{equation}
i = \{\frac{-M}{2}, \frac{-M}{2}+1, \ldots \frac{+M}{2}-1, \frac{+M}{2}\}, j = \{\frac{-M}{2}, \frac{-M}{2}+1, \ldots \frac{+M}{2}-1 , \frac{+M}{2}\}
\end{equation}
\begin{equation}
\psi_x =\frac{1}{P} , \psi_y =\frac{1}{Q}
\end{equation}

{Thus} 
 $\bm{\hat{X}}$ holds the 2D coordinates of all the $M$ $\mathbb{\times}$ $M$ points.
\begin{equation}
S = \mathbb{g}_{ED}(\bm{\hat{X}}, \psi)
\end{equation}

\begin{figure}[H]
\begin{adjustwidth}{-\extralength}{0cm}
\centering
\includegraphics[width=18cm]{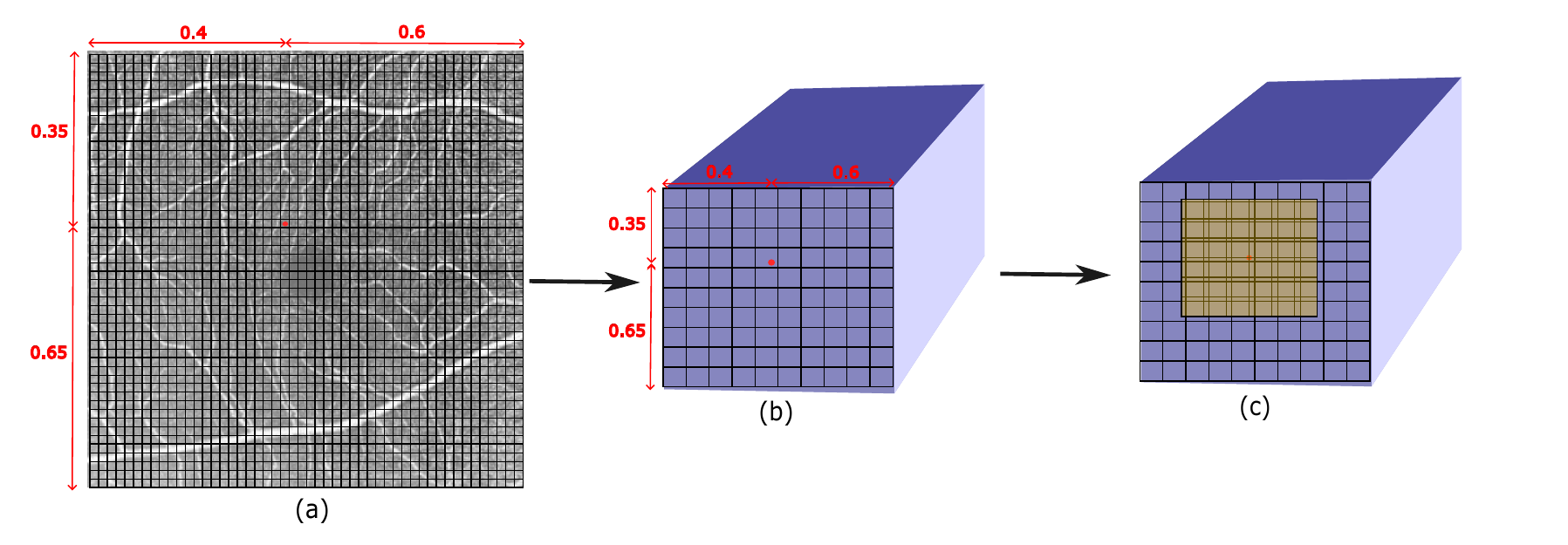}
\end{adjustwidth}
\caption{(\textbf{a}) Given an HR image, a point of interest (red dot) is selected to predict its RGB value. (\textbf{b})~Its corresponding spatially equivalent 2D coordinate is selected from the feature map. (\textbf{c}) Locating the semi-local region (M~=~6) around the calculated 2D coordinate. \label{fig1}}
\end{figure}

{Figure}  \ref{fig1} illustrates how the semi-local region is identified and used to extract the \mbox{$M$ $\mathbb{\times}$ $M$} depth features from the feature map $\psi$. This depiction helps to visualize the steps involved in the feature extraction process.

 \vspace{-12pt}

\subsection{Overlapping Windows\label{sec3.3}}
After extracting the semi-local region $S \in \mathbb{R}^{M \times M \times D}$, our objective is to obtain the RGB value of the center point using this region. To achieve this, we employ a overlapping window-based approach.
We start with four windows, each with a size of $M-1$, positioned at the four corners of $S$. Each window extracts information from its respective region and passes it on to the next subsequent window in the process. With each iteration, the size of the window decreases by 1 until it reaches a final size of $\frac{M}{2}$. This iterative process ensures that information is progressively gathered and refined towards the center point.
This approach allows us to effectively capture and utilize the information from the semi-local region while focusing on the features that are most relevant for determining the RGB value.
\begin{equation}
\Gamma = s_i*w_i
\end{equation}

{In} each iteration $i$, where the window size decreases by 1 for the next step, we utilize weights $w_i$ for combining the features from all four corners. This ensures that the information from each corner is properly incorporated and made available for the subsequent iteration.
In the last step, we take a final window size of 2, but instead of being positioned at the corners as in previous iterations, it is centered around the target point of interest. The features extracted from this final window are then passed through a Multi-Layer Perceptron (MLP) to make the final prediction.

\begin{figure}[H]
\includegraphics[width=10.5 cm]{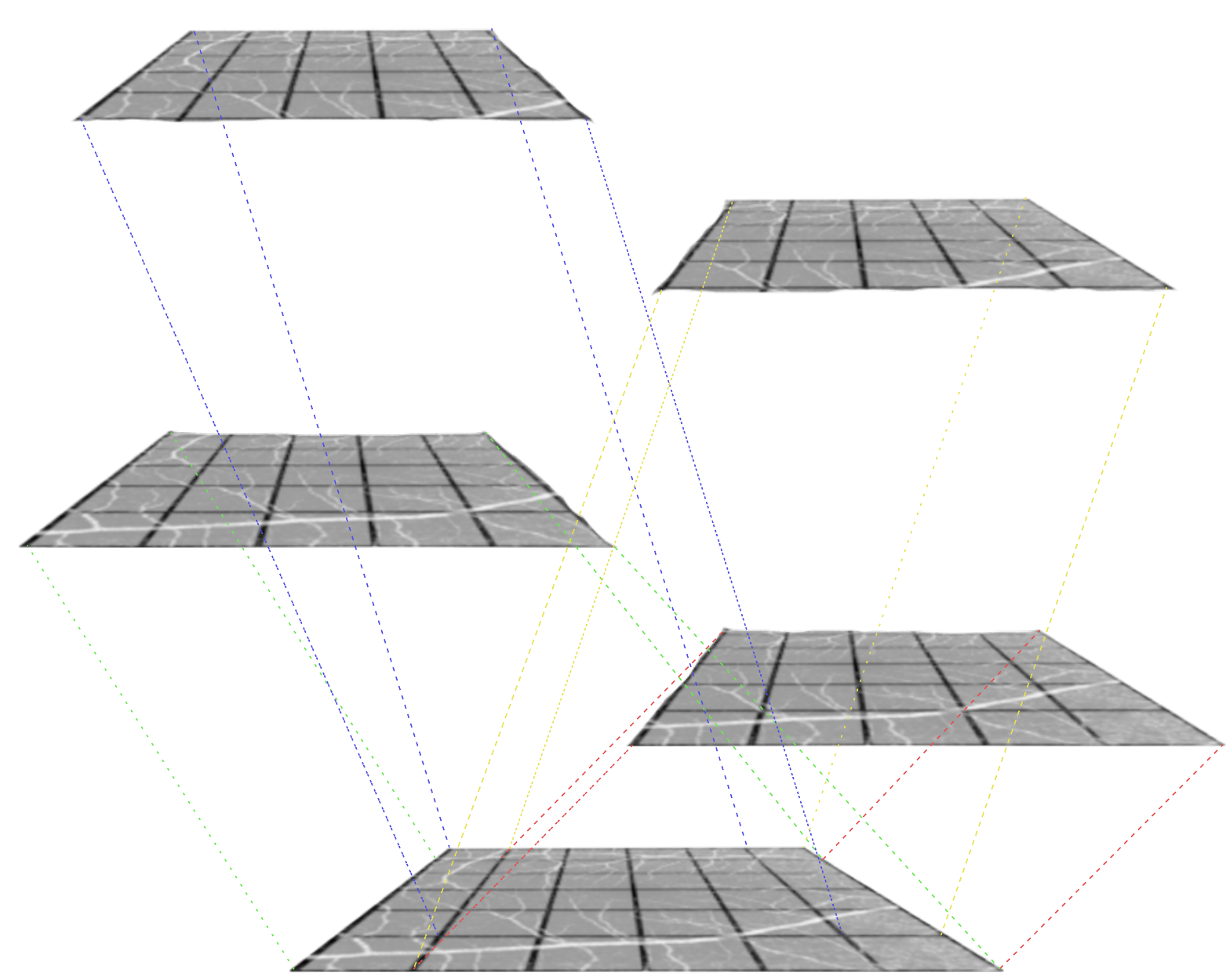}
\caption{The first iteration of overlapping windows, where the window size~=~M~$-$~1 (M~=~6). Assuming the feature map is of negligible depth and four windows are positioned at the four corners of the feature map.\label{fig3}}
\end{figure}

By adapting the window positions and sizes throughout the iterations, we effectively capture and aggregate the relevant information from the semi-local region. This approach allows us to make accurate predictions at the target point, utilizing the combined features from all iterations and the final MLP-based processing. Figure \ref{fig3} shows the working of the overlapping windows.

\unskip


\section{Results and Discussion}
\subsection{Dataset}
We used the OCT500 \cite{Li} dataset and randomly sampled 524 images from it to train our network. It consists of 300 ${3 \times 3}$ OCTA images and 224 ${6 \times 6}$ OCTA images. We use For evaluation, 80 images were selected and we report the results using peak signal-to-noise ratio (PSNR) metric.

\subsection{Implementation Details}

During the training process, we apply downsampling to each image using bicubic interpolation in {PyTorch} 
 \cite{Paszke}. This downsampling is performed by selecting a random factor, which introduces the desired level of degradation to the images. For training, we utilize a batch size of 16 images. From each high-resolution (HR) image, we randomly select 1500 points for which we aim to calculate the RGB values. These points serve as the targets for our network during the optimization process.



To optimize the network, we employ the L1 loss function and use the Adam optimizer~\cite{kingma2014adam}. The learning rate is initialized as $1 \times 10^{-4}$  
 and is decayed by a factor of 0.3 at specific epochs, namely {[40, 60, 70]} 
. We train the network for a total of 100 epochs, allowing it to learn the necessary representations and refine its predictions over time.

Furthermore, each LR image is converted into a feature map of size $48 \times 48$ with a depth of 64 using the EDSR-baseline architecture. This conversion process ensures that the LR images are properly represented and aligned with the architecture used in the training~process.

\subsection{Quantitative Results}
\begin{figure}[H]
\begin{adjustwidth}{-\extralength}{0cm}
\centering
\includegraphics[width=18cm]{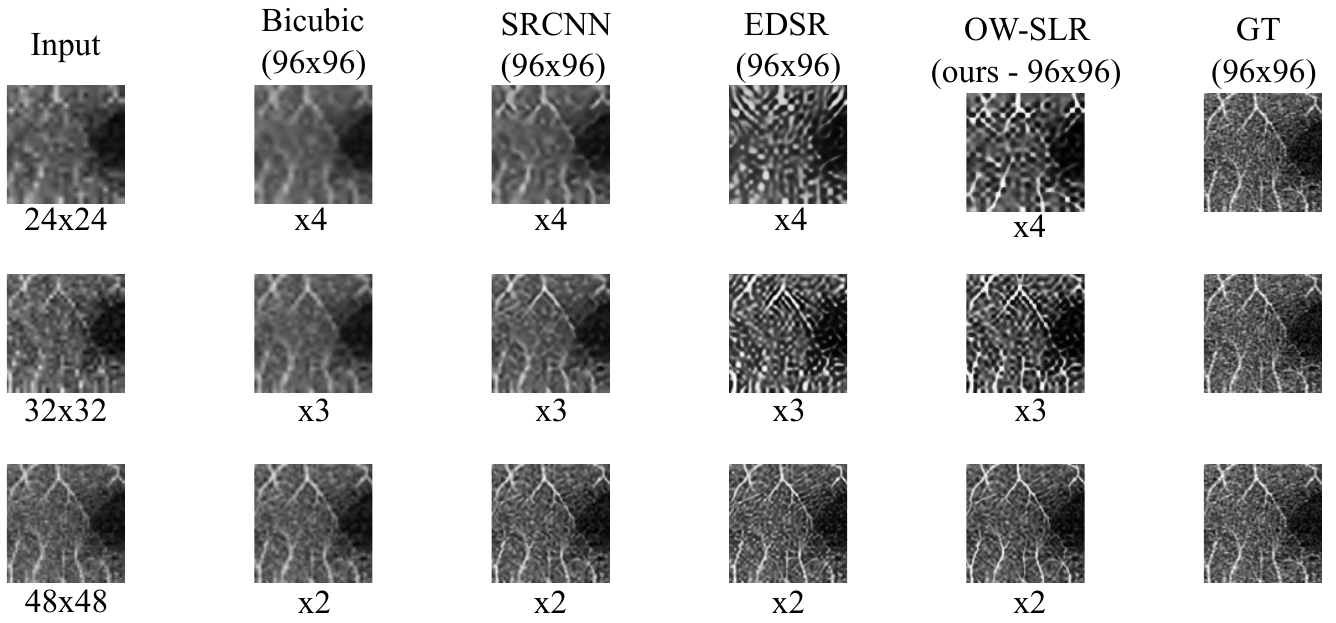}
\end{adjustwidth}
\caption{{A} 
 ${96\times 96}$ patch is taken and its size is reduced to ${24\times 24}$ (first row), ${32 \times 32}$ (second row) and $ {48 \times 48}$ (third row) using bicubic interpolation. Our architecture uses the same set to weights reproduce the given results. However, others require different set of weights for a newer scale to be trained on. The PSNR results of each image are shown in Table \ref{Tab:tab3}. \label{fig4}}
\end{figure}

In Figure \ref{fig4}, we present a comparison of the performance of our proposed OW-SLR method against existing works. The original image patch is first downsampled using bicubic interpolation to a lower resolution. It is evident that there is a significant loss of image quality in the LR patches compared to the ground truth (GT) image. However, our model outperforms the other existing methods, demonstrating a significant improvement when the LR image is extrapolated to a higher scale. The results obtained by our model show better preservation of details and higher fidelity compared to the other approaches when the given image is extrapolated to higher scale. The PSNR results of each image are shown in Table \ref{Tab:tab3}.

\vspace{-12pt}
\begin{table}[H] 
\caption{PSNR result of each of the input images across different methods shown in Figure~\ref{fig4}.}
\begin{tabular}{p{0.15\linewidth}  p{0.15\linewidth}  p{0.17\linewidth}  p{0.17\linewidth} p{0.2\linewidth} }
\toprule
\textbf{Patch Size}& \textbf{Bicubic} & \textbf{SRCNN~\cite{dong2015image}}	 & \textbf{EDSR~\cite{lim2017enhanced}}	 & \textbf{OW-SLR (Ours)} \\
\midrule
24~$\times$~24 & 11.96 & 12.87 & 13.79 & 13.92\\ 
32~$\times$~32  & 14.18 & 15.10  & 16.04 & 16.26 \\ 
48~$\times$~48 & 15.37&  16.89 & 17.66 & 17.98\\
\bottomrule
\end{tabular}
\label{Tab:tab3}
\end{table}\vspace{-3pt}

It is worth noting that our model achieves these results for different scaling factors using the same set of weights trained once. In contrast, the other models would need to be retrained for each new scale to which the LR image is extrapolated. This highlights the versatility and efficiency of our model in handling various scaling factors without the need for additional training.

In Table~\ref{table0}, we provide the upscaling time taken by the proposed model by different factors, while training it just once.

\begin{table}[H] 
\caption{Time taken to extrapolate a ${320\times320}$ image on a single Nvidia Titan V of 12 Gigabyte size. \label{table0}}
\newcolumntype{C}{>{\centering\arraybackslash}X}
\begin{tabularx}{\textwidth}{CC}
\toprule
\textbf{Extrapolation Factor}	& \textbf{Time Taken (In Seconds)}	\\
\midrule
${2\times }$	& 6.48	 \\
${2.4\times }$ &    8.90			\\
${3\times }$ 		&   12.01 \\
${3.9\times }$	&   19.46 \\
${4.5\times }$		&   26.29 \\
${5\times }$	&   33.75 \\
\bottomrule
\end{tabularx}
\end{table}\vspace{-3pt}

In Table \ref{table1}, we present the results of this technique compared to the existing state-of-the-art methods on the OCT500 \cite{Li} dataset. The evaluation metric used in this case is the peak signal-to-noise ratio (PSNR). Our work demonstrates superior performance compared to LIIF, highlighting the effectiveness of considering the semi-local region instead of solely focusing on four specific locations. By incorporating the information from the semi-local region, our approach achieves improved results in terms of PSNR, showcasing the benefits of our methodology for super-resolution tasks.

\begin{table}[H] 
\caption{{PSNR} 
 result on the 300 images from {OCT500} 
 \cite{Li}. \label{table1}}
\newcolumntype{C}{>{\centering\arraybackslash}X}
\begin{tabularx}{\textwidth}{CCC}
\toprule
\textbf{Methods}	& \textbf{PSNR $ \uparrow $}	\\
\midrule
Real-ESRGAN~\cite{wang2018esrgan}	& 15.66		 \\
SRCNN~\cite{dong2015image}		& 16.51			\\
EDSR~\cite{lim2017enhanced}		& 17.49		 \\
LIIF~\cite{chen2021learning}		& 17.60		 \\
OW-SLR (ours)		& \textbf{17.93}	 \\
\bottomrule
\end{tabularx}
\end{table}

\section{Conclusions and Future Work}
OCTA images help us for the diagnosis of retinal diseases. However, due to various reasons like speckle noise, movement of the eye, hardware incapabilities, etc. we lose onto intricate details in the capillaries that play a crucial role for correct diagnosis. We propose this architecture which upscales a given LR image to arbitrary higher dimensions with enhanced image quality. First, we extract the image features using a backbone architecture. We then select a random point in the HR image and calculate its equivalent spatial point in the extracted feature map. We find the semi-local region around this calculated point and pass it through the proposed Overlapping Windows architecture. Finally, an MLP is used to predict the RGB value using the output of the overlapping window architecture. We hope our work will help the people in the medical field in their diagnosis. PSNR 17.93 is achieved for the OCT500 dataset which outperforms the other state-of-the-art work. The technique outperforms the existing methods and allows upscaling images to arbitrary resolution by training the architecture just once.

While effective, it is worth noting that this algorithm does come with a slightly higher computational cost due to its consideration of the semi-local region. There remains potential for further enhancements in both computational efficiency and accuracy while taking the semi-local region into account. This work will provide a stepping stone for future researchers to make strides in this direction.

\vspace{6pt}

\authorcontributions{R.B. has contributed towards implementation of the entire architecture, making the repository publicly available and writing the manuscript, J.J.B. contributed towards knowledge of the data worked upon from the clinical side and article editing, V.L. contributed to advice, suggestions, funding and article editing. All authors have read and agreed to the published version of the manuscript.}

\funding{{This} 
 research was funded by University of Waterloo research grant.}


 
\dataavailability{All generated data are presented in the article body.} 




\conflictsofinterest{The authors declare no conflict of interest.} 

\begin{adjustwidth}{-\extralength}{0cm}

\reftitle{References}

\PublishersNote{}
\end{adjustwidth}
\end{document}